\pdfoutput=1

\documentclass[11pt]{article}

\usepackage[preprint]{acl}

\usepackage{times}
\usepackage{latexsym}

\usepackage[T1]{fontenc}

\usepackage[utf8]{inputenc}

\usepackage{microtype}

\usepackage{inconsolata}

\usepackage{graphicx}

\usepackage{amsmath}
\usepackage{comment}

\title{Probing Internal Representations of Multi-Word Verbs in Large Language Models}

\author{
  \textbf{Hassane Kissane\textsuperscript{1,2}}, 
  \textbf{Achim Schilling\textsuperscript{2,3}}, 
  \textbf{Patrick Krauss\textsuperscript{2,3}} \\
  \textsuperscript{1}Chair of English Philology and Linguistics, University Erlangen-Nuremberg, Germany \\
  \textsuperscript{2}Cognitive Computational Neuroscience Group, University Erlangen-Nuremberg, Germany \\
  \textsuperscript{3}Neuroscience Lab, University Hospital Erlangen, Germany \\
  \thanks{Correspondence: patrick.krauss@fau.de}
}

\begin{document}
\maketitle
\begin{abstract}
This study investigates the internal representations of verb-particle combinations, called multi-word verbs, within transformer-based large language models (LLMs), specifically examining how these models capture lexical and syntactic properties at different neural network layers. Using the BERT architecture, we analyze the representations of its layers for two different verb-particle constructions: phrasal verbs like \textit{give up} and prepositional verbs like \textit{look at}. Our methodology includes training probing classifiers on the model output to classify these categories at both word and sentence levels. The results indicate that the model’s middle layers achieve the highest classification accuracies. To further analyze the nature of these distinctions, we conduct a data separability test using the Generalized Discrimination Value (GDV). While GDV results show weak linear separability between the two verb types, probing classifiers still achieve high accuracy, suggesting that representations of these linguistic categories may be \textit{non-linearly separable}. This aligns with previous research indicating that linguistic distinctions in neural networks are not always encoded in a linearly separable manner. These findings computationally support usage-based claims on the representation of verb-particle constructions and highlight the complex interaction between neural network architectures and linguistic structures.
\end{abstract}

\section{Introduction}
\subsection{The linguistic problem:}
Multi-word verbs or verb-particle combinations are a linguistic category presented in the English language in which the lexical verb is combined with a particle to form an independent unit. It is called a phrasal verb when the lexical verb is combined with an adverbial particle like \textit{work out}. It is a prepositional verb when the verb is combined with a prepositional particle like \textit{rely on} \citep{carter2006cambridge}. Usually, the prepositional verbs are followed by a noun phrase. Rather than the nature of the following particle, there are several differences between phrasal verbs and prepositional verbs.
One main difference between the two categories is the particle placement in phrasal verbs and the foxed order in prepositional verbs. Where in phrasal verbs, the particle can sometimes be separated from the verb and placed after the object. However, the only grammatical form in prepositional verbs is the V+prepostion+object.
\begin{itemize}
    \item \textit{Turn off the light. (phrasal)}
    \item \textit{Turn the light off. (phrasal)}
    \item \textit{Look at the painting. (prepositional)}
    \item \textit{*Look the painting at.(prepositional)}
\end{itemize}

Several studies explored the mental storage of these verb-particle constructions, specifically phrasal verbs, to see in which way they are stored and processed in the brain. For instance \citet{cappell-2010} and further discussed by \citet{Pulvermüller-2013} that phrasal verbs are processed as single lexical units, as evidenced by MEG. However, prepositional verbs remain unexplored, which are still treated similarly to phrasal verbs in terms of both the verb and the particle form a single lexical unit called verb, for example the prepositional verbs \textit{look at}, and the phrasal verb \textit{turn off} \citep{Quirk1985, carter2006cambridge}. From a constructional point of view, \citet{Herbst2008} proposed what is called the valency model for the distinction between phrasal verbs and prepositional verbs, assuming that prepositions function as integral parts of the complement rather than the verb itself in prepositional verbs. This valency-based approach emphasizes the syntactic relationship between the verb and its complements, analyzing prepositional verbs like \textit{look at} as the verb \textit{look} and the complement \textit{at}.

\subsection{probing-based methods for linguistic tasks:}
Probing methods analyze the linguistic properties encoded in the representations of the NLP model. Probes are supervised models trained to predict linguistic properties or other categories, such as parts of speech or word meanings, from model representations such as BERT embeddings \cite{immertreu2024probing, ramezani2024analysis}. These probes have achieved high accuracy on various linguistic tasks, demonstrating their utility in understanding how models encode features such as syntax and semantics \citep{conneau-etal-2018-cram}. The search classifiers are trained on the activations to identify predefined concepts or linguistic properties, such as syntactic tags or semantic meanings, from the model output embeddings \citep{Hupkes2018VisualisationA, sajjad-etal-2022-neuron}. Furthermore, layer-wise analysis \citep{tenney-etal-2019-bert, ramezani2024analysis, krauss2024analyzing, banerjee2025exploring, ramezani2024analysisA} investigates how linguistic knowledge is distributed across the layers of transformer-based models, providing insights into the hierarchical organization of encoded knowledge.

The internal representations of LLMs are frequently analyzed using probing approaches. \citep{tenney-etal-2019-bert} employ probing tasks to investigate the linguistic information that BERT gathers and discover that various layers encode different kinds of linguistic properties. A set of probes is presented by \citep{Tenney2019WhatDY} to examine the representations acquired by contextualized word embeddings and to determine the distribution of syntactic and semantic information among layers.

This study aims to investigate the processing of these two linguistically different multi-word verbs in the internal behaviour of neural language models. Using several interpretability methods such as probing classifiers and data separability methods applied to internal representations of large language models.

\section{Methods}
\subsection{Data:}
The data set consists of sentences containing phrasal and prepositional verbs, divided into training and test sets with labels for each type of verb. The training set includes 1920 examples of phrasal verbs and 2070 of prepositional verbs. The test set contains 522 phrasal verb examples and 623 prepositional verb examples, with a total of 2442 for phrasal verbs and 2693 for prepositional verbs (\ref{tab:verb-dist}). Before using the samples of the dataset as input for the model we applies several cleaning steps in (\ref{tab:preprocessing})

\begin{table}
\centering
\small
\begin{tabular}{lr|lr}
\hline
\textbf{Phrasal} & \textbf{\#} & \textbf{Prepositional} & \textbf{\#} \\
\hline
\multicolumn{4}{c}{\textbf{Training}} \\
\hline
blow\_up & 52 & break\_into & 147 \\
break\_down & 134 & call\_on & 138 \\
close\_down & 54 & come\_across & 168 \\
fill\_up & 54 & do\_without & 76 \\
find\_out & 243 & get\_off & 184 \\
finish\_off & 46 & care\_for & 150 \\
give\_away & 35 & cope\_with & 150 \\
give\_up & 239 & get\_into & 150 \\
hand\_in & 229 & get\_on & 150 \\
hold\_up & 56 & go\_into & 150 \\
look\_up & 67 & lead\_to & 148 \\
put\_off & 57 & listen\_to & 153 \\
shut\_down & 57 & look\_at & 154 \\
throw\_away & 58 & look\_for & 152 \\
turn\_down & 75 & & \\
wake\_up & 31 & & \\
take\_over & 102 & & \\
work\_out & 101 & & \\
sort\_out & 230 & & \\
\hline
Total & 1920 & Total & 2070 \\
\hline
\multicolumn{4}{c}{\textbf{Test}} \\
\hline
take\_up & 100 & depend\_on & 150 \\
carry\_on & 184 & look\_after & 154 \\
bring\_up & 115 & deal\_with & 153 \\
check\_out & 123 & get\_over & 111 \\
& & approve\_of & 55 \\
\hline
Total & 522 & Total & 623 \\
\hline
Grand Total & 2442 & Grand Total & 2693 \\
\hline
\end{tabular}
\caption{Distribution of phrasal and prepositional verbs in training and test sets with their frequencies.}
\label{tab:verb-dist}
\end{table}

\begin{table*}[t]
\centering
\begin{tabular}{|p{7cm}|p{7cm}|}
\hline
\textbf{Character} & \textbf{Pre-processing step} \\
\hline
Punctuations (\texttt{!"\#\$\%\&'()*+,-./:;<=>?@[\textbackslash]\textbackslash \textasciicircum \_ \textbackslash \textasciigrave \{\}|\textbackslash \textasciitilde}) & Removed \\
\hline
Leading and trailing whitespaces & Removed \\
\hline
Extra whitespaces & Replaced with a single space \\
\hline
Uppercase characters & Converted to lowercase \\
\hline
\end{tabular}
\caption{Text pre-processing steps applied to the dataset.}
\label{tab:preprocessing}
\end{table*}

\subsection{Model (Embedding Extraction):}
We use transformer-based model \citep{Vaswani2017AttentionIA} BERT \citep{devlin-etal-2019-bert} as the feature extraction model for generating contextual embeddings. Specifically, we use the \textit{bert-base-uncased} version, consisting of 12 layers, each producing 768-dimensional contextual embeddings for input tokens. For each sample, we extract embeddings at two levels:

\textbf{Token-Level Embeddings:} For verb-specific analysis, we extract the embedding corresponding to the main token of the verb (e.g., \textit{give} in \textit{give up}). These embeddings focus on the localized representation of the verb within the sentence.

\textbf{Sentence-Level Embeddings:} To capture the entire context of the sentence, we compute the average of all token embeddings in the sentence. This approach aggregates information across all tokens, providing a representation of the sentence without relying solely on the [CLS] token embedding.

\subsection{Classification Models:}

\textbf{Logistic Regression (LR):} Logistic regression is a linear model used in modeling the probabilities of possible outcomes given an input variable.

\textbf{Support Vector Machines (SVM):}, is the best working algorithm with smaller datasets than large ones, by optimizing data transformations based on predefined classes or outputs. It is based on the principle of Structural Risk Minimization from the Statistical Learning Theory \citep{SVM}. In their fundamental form, SVMs learn linear discrimination that separates positive examples from negative ones with a maximum margin. This margin, defined by the distance of the hyperplane to the nearest positive and negative examples, has proven to have good properties in terms of generalization bounds for the induced classifiers.

\subsection{Generalized Discrimination Value (GDV):}

We used the GDV to calculate cluster separability as published and explained in detail in \citep{schilling2021quantifying}. Briefly, we consider $N$ points $\mathbf{x_{n=1..N}}=(x_{n,1},\cdots,x_{n,D})$, distributed within $D$-dimensional space. A label $l_n$ assigns each point to one of $L$ distinct classes $C_{l=1..L}$. In order to become invariant against scaling and translation, each dimension is separately z-scored and, for later convenience, multiplied with $\frac{1}{2}$:
\begin{align}
s_{n,d}=\frac{1}{2}\cdot\frac{x_{n,d}-\mu_d}{\sigma_d}.
\end{align}
Here, $\mu_d=\frac{1}{N}\sum_{n=1}^{N}x_{n,d}\;$ denotes the mean, and $\sigma_d=\sqrt{\frac{1}{N}\sum_{n=1}^{N}(x_{n,d}-\mu_d)^2}$ the standard deviation of dimension $d$.
Based on the re-scaled data points $\mathbf{s_n}=(s_{n,1},\cdots,s_{n,D})$, we calculate the {\em mean intra-class distances} for each class $C_l$ 
\begin{align}
\bar{d}(C_l)=\frac{2}{N_l (N_l\!-\!1)}\sum_{i=1}^{N_l-1}\sum_{j=i+1}^{N_l}{d(\textbf{s}_{i}^{(l)},\textbf{s}_{j}^{(l)})},
\end{align}
and the {\em mean inter-class distances} for each pair of classes $C_l$ and $C_m$
\begin{align}
\bar{d}(C_l,C_m)=\frac{1}{N_l  N_m}\sum_{i=1}^{N_l}\sum_{j=1}^{N_m}{d(\textbf{s}_{i}^{(l)},\textbf{s}_{j}^{(m)})}.
\end{align}
Here, $N_k$ is the number of points in class $k$, and $\textbf{s}_{i}^{(k)}$ is the $i^{th}$ point of class $k$.
The quantity $d(\textbf{a},\textbf{b})$ is the euclidean distance between $\textbf{a}$ and $\textbf{b}$. Finally, the Generalized Discrimination Value (GDV) is calculated from the mean intra-class distances

\begin{align}
\langle\bar{d}(C_l)\rangle=\frac{1}{L}\sum_{l=1}^L{\bar{d}(C_l)}
\end{align}

and the mean inter-class distances 

\begin{align}
\langle\bar{d}(C_l,C_m)\rangle=\frac{2}{L(L\!-\!1)}\sum_{l=1}^{L-1}\sum_{m=l+1}^{L}\bar{d}(C_l,C_m)
\end{align}

as follows:

\begin{align}
\mbox{GDV}=\frac{1}{\sqrt{D}}\left[ \langle\bar{d}(C_l)\rangle - \langle\bar{d}(C_l,C_m)\rangle \right]
 \label{GDVEq}
\end{align}

\noindent whereas the factor $\frac{1}{\sqrt{D}}$ is introduced for dimensionality invariance of the GDV with $D$ as the number of dimensions.

\vspace{0.2cm}\noindent Note that the GDV is invariant with respect to a global scaling or shifting of the data (due to the z-scoring), and also invariant concerning a permutation of the components in the $N$-dimensional data vectors (because the euclidean distance measure has this symmetry). The GDV is zero for completely overlapping, non-separated clusters, and it becomes more negative as the separation increases. A GDV of -1 signifies already a very strong separation.

\section{Results}
\textbf{Token-based classification:}

The results of the lexical verb token classification task using logistic regression and linear SVM have shown distinct trends across the 12 layers of the
BERT model. For the Logistic Regression classifier, accuracy starts at 0.87 in the input layer, remains stable around 0.84 - 0.80 through layers 2 to 4, and then increases significantly, reaching 0.99 in layer 6 before slightly decreasing in the late layers of the model. Similarly, the linear SVM classifier achieves an accuracy of 0.63 in the input layer. Then
0.84, through layers 2 to 4. To start improving from layer 5 onward, reaching its highest accuracy of 0.99 at layer 8. Both classifiers show the best accuracy in the middle layers (layers 6–9, suggesting that these layers encode the most significant linguistic features to distinguish between phrasal verbs and prepositional verbs. Therefore, the accuracies decrease slightly in the late layers, which indicates a shift towards task-specific representations less suited for this classification task.

\textbf{Sentence-Based Classification:}

For the sentence-based classification task, both Logistic Regression and Linear SVM show distinct trends across the 12 BERT layers. However, the accuracies in the token-based classification were higher than those based on sentence embeddings. The Logistic Regression classifier begins with an accuracy of 0.80 in the input layer and improves to 0.85 in layers 6 and 7. Then, accuracy decreases in the late layers, dropping to 0.69 in layers 11 and 12. Similarly, the linear SVM classifier starts with an accuracy of 0.77 and 0.76 in the input and first layer respectively, to 0.84 in layer 6, then decreases to the lowest accuracy of 0.66 in the final layer of the model. With these results, it is suggested that the middle layers (layers 5–7) of the model are the best to capture linguistic information at sentence-level representations to distinguish phrasal verbs and prepositional verbs, while the higher layers, likely focused on task-specific semantics, encode features less suited to these properties predictions.

\begin{figure}[t]
  \centering
  \includegraphics[width=\columnwidth]{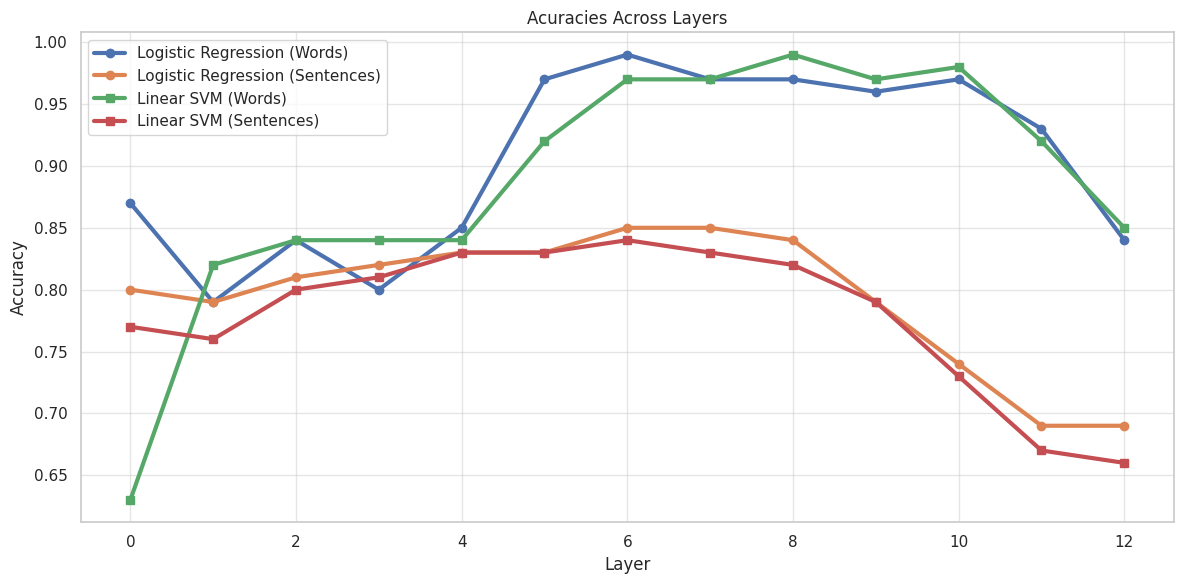} 
  \caption{Classification accuracies}
  \label{fig:sample_small1}
\end{figure}

\textbf{GDV Values for Data Separability:}

The GDV calculations for both token-based and sentence-based embeddings has shown non-strong separation between between the phrasal and prepositional verbs across BERT layers. For the token-based embeddings, GDV values start at equivalent of 0.00 in the input layer which is responsible for converting tokens into dense vector representations before they are processed by the transformer layers. Then, the GDV has shown an improvement (less negative) across the layers, reaching their strongest separability at layers 3 and 4 with a value of -0.049 and -0.048 respectively. This improvement indicates that BERT's middle layers may encode more discriminative features for distinguishing between the two verb types in word embeddings.

After we ran a normality test, we found that the data was not normally distributed. therefore we test the correlation between the classifiers and the GDV values. The Spearman's rank correlation analysis showed no statistically significant correlation between the variables. For words, the correlation between Logistic Regression and GDV was \( r_s\) = 0.32, \( p\) = 0.285, and between Linear SVM and GDV, \( r_s\) = 0.26, \( p\) = 0.383. For sentences, the correlation between Logistic Regression and GDV was \( r_s\) = -0.44, \( p\) = 0.128, while Linear SVM and GDV showed a negative correlation of \( r_s\) = -0.52, \( p\) = 0.069, approaching significance. Overall, no strong or significant associations were observed.

\begin{figure}[t]
  \centering
  \includegraphics[width=\columnwidth]{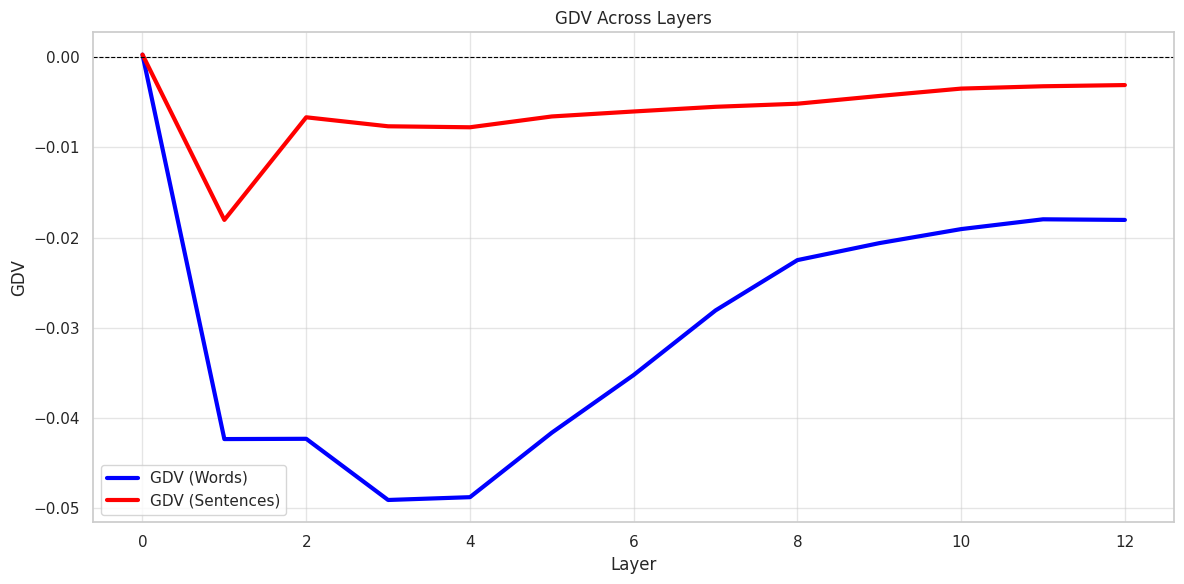} 
  \caption{GDV values for data separability between the two multi-word verbs classes across BERT layers}
  \label{fig:sample_small2}
\end{figure}

\begin{figure*}[ht]
    \centering
    \includegraphics[width=\textwidth]{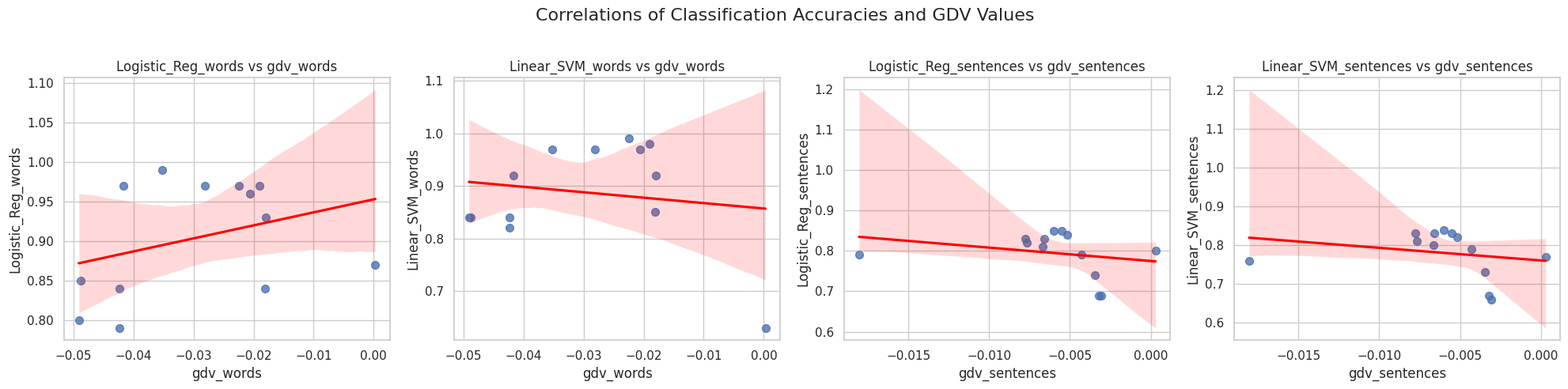}
    \caption{Correlations of Classification Accuracies and GDV Values.}
    \label{fig:correlations}
\end{figure*}

\section{Discussion}

Our findings indicate that while probing classifiers and GDV provide some investigations into how BERT encodes differences between linguistic categories, they may not fully capture the complexities of linguistic representations. Particularly, when distinguishing between phrasal and prepositional verbs based on token-level embeddings. As proposed by \citet{goldberg1995constructions}, lexico-semantic elements convey a portion of linguistic information, but they do not embody all structural and functional aspects present in a text. This limitation is particularly relevant in the case of the investigated cases in the study, where distinctions often emerge from interactions between lexical, syntactic, and semantic factors rather than being determined by individual token representations.

This perspective aligns with the constructionist approach to language processing \citep{Madabushi2020}, which challenges the traditional separation of lexical and grammatical elements. Instead, it proposes a continuum of constructions—where linguistic representations arise from learned pairings of form and meaning rather than being strictly lexical or grammatical only. From this point, phrasal and prepositional verbs might be better understood as integrated constructions, rather than purely lexical or syntactic units. Consequently, probing classifiers, which primarily capture lexical or semantic properties in token-based classification tasks, may fail to fully account for the grammatical and contextual information that shapes the representation of these constructions. This is evident in the mismatch between classification accuracies and GDV values, suggesting that different methods may capture other dimensions of representation.

Several studies have discussed the limitations of probing classifiers \citep{Blinkov-2022, sajjad-etal-2022-neuron}. One major limitation is the disconnect between probing accuracy and the original model’s internal processing. While probing classifiers can detect correlations between model embeddings and linguistic features, they do not necessarily indicate whether the model actively uses these properties for linguistic processing. This limitation is apparent in our findings: while probing classifiers achieved high accuracies, GDV analysis showed weak linear separability between phrasal and prepositional verbs, as indicated by the lack of significant correlation between classification accuracies and GDV values.

This observed Disagreement between classifier accuracies and GDV values aligns with previous research suggesting that internal representations in neural networks and large language models are not necessarily linearly separable \citep{hewitt-liang-2019-designing, kissane2024analysis, zhang-bowman-2018-language, banerjee2025exploring, hildebrandt2025refusal, krauss2024analyzing, ramezani2024analysis}. Since GDV measures linear separability, it does not capture non-linearly structured representations. In contrast, probing classifiers can still detect non-linearly separable distinctions, allowing them to identify linguistic categories that may be encoded in high-dimensional space. Therefore, the low GDV scores do not suggest that BERT fails to encode multi-word verb distinctions, but rather that these representations may require non-linear transformations to be fully distinguished. This Point up the need for comprehensive analytical methods when investigating how LLMs structure linguistic knowledge and suggests that linear separability should not be the only one criterion for assessing learned representations.

\section*{Acknowledgments}

This work was funded by the Deutsche Forschungsgemeinschaft (DFG, German Research Foundation): KR\,5148/3-1 (project number 510395418), KR\,5148/5-1 (project number 542747151), and GRK\,2839 (project number 468527017) to PK, and grant SCHI\,1482/3-1 (project number 451810794) to AS.

\end{document}